# Bridging the Gap between Artificial Intelligence and Artificial General Intelligence: A Ten Commandment Framework for Human-Like Intelligence

Ananta Nair[1,2] and Farnoush Banaei-Kashani[1]

1. University of Colorado, Denver, 2. Dell Technologies Inc

## Abstract

The field of artificial intelligence has seen explosive growth and exponential success. The last phase of development showcased deep learnings ability to solve a variety of difficult problems across a multitude of domains. Many of these networks met and exceeded human benchmarks by becoming experts in the domains in which they are trained. Though the successes of artificial intelligence have begun to overshadow its failures, there is still much that separates current artificial intelligence tools from becoming the exceptional general learners that humans are. In this paper, we identify the ten commandments upon which human intelligence is systematically and hierarchically built. We believe these commandments work collectively to serve as the essential ingredients that lead to the emergence of higher-order cognition and intelligence. This paper discusses a computational framework that could house these ten commandments, and suggests new architectural modifications that could lead to the development of smarter, more explainable, and generalizable artificial systems inspired by a neuromorphic approach.

## Introduction

Though the concept of artificial intelligence (AI) may seem like a futuristic prospect, the desire to create man-made intelligence has long been a human yearing. Dating back to approximately 700 BCE is the ancient Greek myth of Talos, the first robot. Talos was created by the god Hephaestus to serve as guard of Crete and hurl boulders at incoming vessels. One day a ship approached the island, which, unknowing to the behemoth automatron, would serve as his greatest challenge. In an attempt to escape the machine, Medea, a sorceress onboard, constructed an ingenious plan and offered the robot a bargain; granting him eternal life in return for removing his only bolt.

Surprisingly, this offer resonated with Talos, who had not yet come to grips with his own nature, nor understood his longings for human desires such as immortality. Though the tale ended tragically for the giant, it does speak to the desires and fears humans have long had of creating intelligence and the blurred line between man and machine, a challenge that is becoming increasingly prominent today.

From mechanical toys of the ancient world to the looming dystopian apocalyptic scenarios of science fiction, the creation and progression of AI has long been on the human mind. Though many would argue the field has seen many bursts and busts, the last decade has resulted in the most remarkable progress till date. Largely these advances have come from deep neural networks that have become experts in areas such as vision, natural language processing, and reinforcement learning (Brown et al., 2020, Ramesh et al., 2022, Chowdhery et al., 2022, Schrittwieser et al., 2020, Ye et al., 2021, Baker et al., 2019, Arnab et al., 2021). Their breadth of application is expansive and the commercial and practical deployments being undertaken seem almost endless. Applications such as facial recognition ((Balaban et al., 2015), have found common place on our phones and in our daily lives; virtual assistants like Alexa are a significant step up from IBM's Shoebox (Soofastaei, 2021), that could recognize only sixteen words and digits; online translators are capable of accurately translating between any two or multiple languages (Fan et al., 2021), and reinforcement learning agents have achieved human and superhuman performance on a range of complex tasks such as board and video games (Schrittwieser et al., 2020, Silver et al., 2018, Berner et al., 2019, Vinyals et al., 2019). Even in academia, these tools have resulted in significant breakthroughs in a diverse set of endeavors, including weather prediction, protein unfolding, mental health, medicine, robotics, and astrophysics (Ravuri et al., 2021, AlQuraishi, 2019, Su et. al, 2020, Lusk et al., 2021, Gillenwater et al., 2021, Pierson & Gashler, 2017, Huerta et al., 2019, George & Huerta, 2018).

Inarguably, each of these undertakings deserve to be commended for its unique accomplishments. However, there still remains much work to be done by the field to achieve general intelligence akin to the natural world. Deep networks in collaboration with the advancement of GPUs, have accelerated data-processing to master pattern recognition and other statistical based learning, supervised and rule based learning, unsupervised and training-test generalization, and reinforcement and multi-agent learning (Zhu, 2005, Caruana & Niculescu-Mizil, 2006, Sutton & Barto, 2018, O'Reilly et al., 2021, Mollick et al., 2020, Baker et al., 2019, Stooke et al., 2021). These ingenious mathematical algorithms and data-intensive processing techniques lead to great success in ideal conditions by converting a sparse problem space into a dense sampling whose distribution can be integrated over or overfit on. These networks can even further transfer to out-of-distribution learning, doing so by integrating over both training and testing sets (Vogelstein et al., 2022). However, a common side-effect and growing problem of this type of approach has been the creation of data-sensitive black boxes that struggle to generalize outside of their well defined parameterizations (Lake et al., 2017, Geirhos et al., 2020). These challenges make current AI tools great optimizers but they still fall short of traditional intelligence.

In comparison, natural intelligence is an exceptional example of evolutionary engineering, with the human brain regarded as the premier example. Given the latest successes but large limitations of deep networks, the field of AI is now more than ever drawing comparisons and parallels between what neural networks can do and what natural intelligence is capable of (Vogelstein et al., 2022, Silver et al., 2021, Richards et al., 2019). The brain, unlike AI, does not demonstrate or strive for exceptional performance at every task it attempts in a hopeless effort to maximize its reward function. Instead, it creates exceptional universal learners that are capable of generalizing their abstract representations and skills to learn any task quickly and easily. This methodology does not lead to exceptional general performance across all tasks, but instead behavior is adjusted through goal-driven learning to create a system that only excels at high value objectives whereas other tasks have more moderate probabilities of success. This type of learning optimizes for specific performance, while leaving enough computational resources to adequately perform all tasks. For example: athletes and soldiers may excel at tasks that require scene integration, strategy, and physical or reflex capabilities, however scientists would not need to excel in these domains and may instead showcase exceptional performance on logic and reasoning, or artists on fine-tuned motor movements. Though each profession can learn the other's skill set, humans prioritize their performance based on high value objectives that align with their goals.

We believe this methodology of maximizing an architectures learning based on creating general goal-directed learners rather than overtrained task specific agents is the key to the emergence of more intelligent systems. In this paper, we use the brain as inspiration, to identify the key properties that we believe are essential to higher order cognition. The next section of the introduction teases apart components of brain function that could address notable limitations seen in current AI tools. Whereas, the last section of the introduction provides an overarching summary of the framework. Next, section two addresses the key properties of brain function that we believe lead to intelligence. These are communicated as the ten commandments, and we believe intelligence emerges not from an individual commandment but rather an all encompassing system whose collective sum is stronger than its part. Lastly, section three addresses how these commandments can be proposed into a framework for the development of AI systems that transition from statistical superiority to general intelligence. Though this paper is largely addressing the assembly of components that may lead to autonomous AI akin to natural intelligence, we believe that the individual commandments in themselves can be beneficial to improve a myriad of AI systems and tools where the end goal is not an all encompassing system. Through the presentation of the commandments and the framework, we believe models can be scaled up or down as needed.

## 1.1 Teasing the Brain Apart; The Paradox of Intelligence

Numerous deep learning models have surpassed human performance on tasks on which they have been intensively trained (Schrittwieser et al., 2020, Silver et al., 2018, Berner et al., 2019, Vinyals et al., 2019). This has led some to even argue that these models are akin to humans in the manner in which they abstract the world and perform a strategic look ahead (Cross et al., 2021, Buckner, 2018). Though this may or may not be true, it can be agreed upon that there exists a notable list of sizable limitations that prevent AI tools from demonstrating the fluid intelligence that makes humans the exceptional general learners they are. Though there exist numerous issues in AI, we believe the most notable limitations are; 1) long training times (Schrittwieser et al., 2020, Chowdhery et al., 2022, Berner et al., 2019); 2) the impotence to chart a novel path to better guide learning; and 3) the inability to generalize to new or increasingly complex, uncertain, and rapidly changing domains (Poggio et al., 2019, Geirhos et al., 2020). The scheme of learning and retaining task-specific data which makes deep networks so successful, also limits it by forcing networks to use a computationally intensive data-driven tabula rasa approach rather than allow models to build upon what they know. Problems with this rigidity has not only lead to long training times but also in complex domains has resulted in the daunting investigation of the trustworthiness and reliability of these networks and the understanding they truly have of the task which they're undertaking (Geirhos et al. 2020, Hubinger et al., 2019, Koch et al., 2021).

On the other hand, it is believed natural intelligence, be it humans or animals, are either born with or soon after develop innate knowledge which they can use to build increasingly complex hierarchical representations of their environment that unfold over time. Though what this knowledge is, and how it has been evolutionary encoded is debated, it suggests that natural intelligence does not abide by either a blank slate or a tabula rasa approach (Wellman & Gelman, 1992, Lake et al., 2017, Velickovic et al., 2021, Silva & Gombolay, 2021). Furthermore, it is becoming increasingly well accepted that the brain takes in the world by breaking it down into its smallest components, even though neuroscientists are unsure how inputs are processed to create internal models. All that is known, is that these smallest components or concepts as they are often called (van Kesteren et al., 2012; Gilboa & Marlatte, 2017; van Kesteren & Meeter, 2020), are hierarchically combined with increasing complexity to generate an internal model of the environment. It has been widely hypothesized that the brain is able to do so by organizing information into global gradients of abstraction. These gradients in accordance with a relational memory system inclusive of processes, such as maintenance, gating, reinforcement learning, memory, etc., continually update, store, recombine, and recall information that unfolds and is strengthened over time to form generalized structural knowledge (Whittington et al., 2019).

A secondary backbone of intelligence is the ability to contextualize the world into an associative framework of increasing complexity and actionable goals and subgoals (O'Reilly et al., 2014, O'Reilly, R. C. et al., 1999, Reynolds & O'Reilly, 2009, O'Reilly, 2020). As argued in Hubinger et al., 2019 and Koch et al., 2021, current deep networks establish objectives to assign to an optimizer but then utilize a different model tasked with carrying out actions. This results in an optimizer with an assigned objective, that then in turn optimizes a model that can act in the real world. This type of architecture not only leads to problems in matching goal alignment between creator and creation but also results in problems in trust and explainability. For example: the authors placed an agent in an environment where it had to find and collect keys that it must use to open a chest and gain reward. The difference between the training and testing environments was the frequency of the objects, with the training set having more chests than keys and the testing having more keys than chests. It was found that this simple difference in environment was strong enough to force the agent to learn a completely different strategy than intended, i.e. finding keys is more valuable than opening chests. This result occurred unsurprisingly as the agent valued keys as a terminal goal and not a sub-goal. Thus, in testing, the agent not only collected more keys than it could use but also repeatedly circled the area of inventory keys displayed on the screen. These types of unintended strategies commonly seen by deep networks when deployed in an environment outside of its training set have resulted in an ever-growing concern and increasing skepticism of the outputs of deep networks.

Humans, and even children on the other hand could solve this task with much ease, as they use a combination of strategies to determine their actions. A child, when tasked with either playing or watching another person play a video game, is capable of inferring what the objectives of the game are and what actions can be considered good (rewarding) or bad (punishable) based on only a few learning instances. We believe that it is likely that the level of abstractions young children have are not very different from that of a well-trained neural network, i.e. the complexity and data points constraining and defining the abstraction are likely of a similar degree. For example: a child will be able to successfully identify animals they have interacted with however would struggle with ambiguous examples of a dog or cat just as a convolutional neural network would. However, the reason a child could solve the tasks a deep network fails on is due to a mixture of the following techniques. Though initially the child might have a high rate of error, they are quickly able to learn by (1) utilizing a combination of learning techniques such as instructional, supervised, self-supervised, unsupervised, predictive, and trial and error learning guided by a reinforcement learning signal to form flexible representations of their environment that achieves the same outcome, (2) utilizing similarity and dissimilarity as a benchmark to compare and group new representations based on what they already know, (3) asking probing questions such as how, what, and why to determine state transitions, (4) establishing and determining goals and objectives that need to be met by the agent, (5) being able to go offline to combine representations in new and novel ways, and (6) being able to establish causal relationships between representations.

We believe each of these learning strategies that emerge in natural intelligence allows for children to take the abstractions that they know and build complex models as adults. To root this in an example, let us explore how a child would learn to make coffee, or the coffee making task as it is often called in robotics (Tsai et al., 2010). Children develop flexible motor movements and an understanding of their environments by continuously interacting with their world. They do not learn from one type of training such as supervised or unsupervised but utilize a combination of different training techniques and steps to achieve the same outcome. This could include learning from instructions, watching someone else, as well as using trial and error techniques. Furthermore, the representations formed by utilizing multiple but interconnected modalities such as vision, auditory, somatosensory etc. helps constrain categories and form richer representations. In children, the representation of a coffee pot would include the visual characterization, the sound and feel of its various components such as glass and plastic, as well as the shape and detachable top affording it the capability of holding liquids. We believe both the flexibility and the richness of multi-modal characterization of representations is essential for allowing higher level cognitive regions to build complex representations and plans for the world.

Through this process of learning to achieve a goal, the child would ask questions internally or externally to reason why and how specific actions need to be performed or lead to specific outcomes. This could include reasoning if a particular object is large enough to hold coffee or water as well as what combination of steps should be used, load coffee or water first. Children further contrast new representations of the environment with past representations of what they've experienced. This could suggest scenarios such as contrasting the coffee pot with other liquid holding containers, such as water bottles and jugs, or contrasting it with past experiences with machines they may have seen or interacted with. The child would use these representations along with instructed or self-imposed action-outcome pairings to learn rules that can formulate into better and faster actionable plans. An example of such a rule could include the coffee is placed in the smaller container, and the water in the larger container or five scoops of coffee are needed for twelve cups. The learning that occurs on each step of the task from the real world, we believe, would result in a much more tangible and potent reward signal than a network could receive. For example, if the child would omit the step of adding coffee or adds too much coffee to the machine, they would receive a much more tangible reward or punishment signal than those received by traditional deep networks, promoting faster learning. The learning could be used both online to form cause and effect relationships between actions or outcomes, or offline, during thinking to simulate and streamline actions for faster or better results in the future.

On a neurological note, we believe an additional virtue that maximizes learning, and the ability to create abstractions that can generalize or transfer across task domains in the brain is the use of specialized experts. These architectures, which are structurally designed, take in specific inputs for a particular type of processing and produce functionally outputs that establish associations between structural experts. Structural connectivity, much as the name suggests, is the connectivity that arises from specific structures in the brain, such as the white matter fiber connectivity between gray matter regions (Hagmann et al., 2008; Iturria-Medina et al., 2008; Gong et al., 2009; Abdelnour, Dayan, Devinsky et al., 2018). Functional connectivity, on the contrary, is concerned with relationships between brain regions, and typically does not rely upon assumptions about the underlying biology. The functional signal refers to strength and activation of pairs of brain regions over time. Regions are shown to have functional connectivity if there is a statistical relationship between their activity over recorded time. It can be interpreted as the temporal correlations and undirected association between two or more neurophysiological time series (such as those obtained from fMRI or EEG) (Chang & Glover, 2010; Abdelnour, Dayan, Devinsky et al., 2018). These structural and functional representations can be hierarchically processed with increasing complexity across levels defined by specific constraints as discussed below. Structural and functional connectivity play distinct but interacting roles to achieve abstraction and subsequently higher-level cognition within the brain. This method of compartmentalizing knowledge for training, we believe, serves as a key optimizer for performance and generalization.

## 1.2 Assembling the Brain; The Roadmap to a Framework

As indicated above, in addition to identifying the properties key to natural intelligence, we propose a framework which would serve as a means to house them, i.e., the ten commandments. A defining feature of this framework is that it does not draw its strength from an individual commandment, but rather from a system that combines all the components to emerge stronger than its parts. At its essence, we believe the brain operates on fundamental principles that can be isolated and computed. However, it's the intricate interaction of these principles that leads to intelligence, which is difficult to isolate and define. The aim of this paper serves to identify these fundamental principles and suggest a computational means for executing intelligence to inform the next wave of both intelligent and autonomous AI.

We imitate optimization and cognitive control in the brain by organizing information into two hierarchical loops of experts: an inner and an outer. This is inspired heavily by the global workspace theory (Baars 1993, 1994, 1997, 2002, & 2005) and hierarchical theories of cognitive control (Cushman & Morris, 2015, O'Reilly et al., 2020). This hierarchy drives behavior to emerge from a shared system where the inner loop is associated with processing and accomplishing a specific task. This could include learning how to make coffee or other tasks such as navigation, language or translation, and informational analysis. Whereas the outer loop performs an integrative function of assembling information from the inner loop to achieve more complex and longer time-scale cognition (O'Reilly et al., 2020). For instance, this could include streamlining the process of making coffee to the most efficient strategy, transferring previous knowledge for making coffee to a new machine, or even previous knowledge of making coffee to other beverages like tea or hot chocolate. Similarly, across other tasks it could include navigating to different goals, determining similarities between languages it knows and doesn't know to understand words, and even going beyond statistical determinism to provide contextualization and personalization to informational analysis.

For simplicity, the framework as shown in Figure 1, is divided out into three layers with a set number of experts. Though specific numbers were chosen for illustrative purposes, we believe that there can be $n$ number of layers with either $n$ or $m$ number of experts. The only condition is that the inner and outer loop must imitate one another with the same number of layers and the same number of experts per layer. This is due to the outer loop being a mimic of the inner loop, that integrates only the most salient or reward gated information from the inner loop. In the brain, as there are no duplicative regions, the same brain regions perform the role of both loops. However, for better explainability and transparency, we duplicated the loops into two separate entities. Gated representations such as those leading to reward or punishment as well as salient events or objects in the environment recorded by the inner loop are passed to the respective outer loop experts and layers. This happens for all tasks, so that the outer loop can integrate and consolidate the most relevant representations. The outer loop's processing involves refining representations to develop better plans, recombining representations to create new and novel actions, and exerting top-down influence on the inner loop to improve real-time processing.

For both loops, each layer receives from a number of experts which are special architectures with attributes that maximize performance for a particular trait (for example: vision, auditory, somatosensory etc.). The information from the structural expert models is processed and outputted in a common language that can be integrated and constrained via lateral connections in the layer to promote a multi-modal representation. For example: constraining the image of a coffee machine, with the sound of brewing coffee, to form a more complex category of coffee machine. This same processing could apply to a variety of different domains, including the cat vs dog classification task, whereby the image of a dog or cat is constrained with its sound, and somatosensory profile to form a more complex category for recognizing the animal. The final common language output is the functional representation for the category and encompasses the final core constrained information across multiple structural experts. As indicated above this could include a complex multimodal representation of the category determined by the laterally connected experts. The functional information output from layer one is fed into the structural experts of the second layer, who in turn use the information to create more complex functional representations to input to layer three. This process can further be repeated at each increasing level of the hierarchy. For instance processing at level, one could result in the formation of multimodal representations, whereas processing at level two could result in scene integration and proposing of actions, and lastly processing at level three would involve evaluating actions based on predictions as well as motor execution.

Both the inner and outer loops operate in this hierarchical manner, with the only two differences being 1) the level of abstraction formed and the complexity of the processing taking place; and 2) only the inner loop interacts with the real world, the outer loop can only do so indirectly through top-down influence of the inner loop. When a particular action-outcome pairing or series of actions is rewarded, a salient or unexpected event occurs, or a large punishment is given, the system learns by gating information that trains up good representations and trains down bad representations. Each of these gated inputs from the various inner loop layers are then passed to the corresponding outer loop layers to retain, refine, and recombine representations in a slower more integrated manner. This process allows for the grouping of actions to create more flexible generalizable representations that can be applied to the same or similar task in the future. The outer loop can then use these representations to exert top-down influence on the inner loop to act faster or use a new combination of actions. For instance, if after trial and error a series of actions successfully brews coffee, all the representations involved in that process would be gated into the outer loop. However, as the inner loop repeats the task, the outer loop would learn to streamline all the gated representations to advise the inner loop to only take the set of actions that lead to reward the fastest.

Furthermore, to better direct behavior, the framework's outputs are structured to be goal directed, depicted as a global $F(X)$ function where $X$ is being maximized for a specific goal. To further tune the system, other constraints such as attention, maintenance, and gating are also used. The system creates multiple goals each optimized to a specific behavioral strategy (such as making coffee as described above, obtaining food or battery charge when hungry or resources are needed, and shelter when escaping weather elements). Each of these goals maximize a series of actions that lead to reward or punishment by using attention to attend to the most relevant information, and gating and maintenance to encourage the system to retain that information. However, it should be noted, to not give the optimization of goals and the outer loop complete control of the system, constraints such as attention, can redirect the model to focus on more pressing goals that are time-sensitive or salient environmental events that can threaten the overall health of the system. Such as the presence of a predator when searching for food, or the presence of a fire in the kitchen when making coffee.

Lastly, the learning, retention, refining, and recombination of information which can be referred to as memory is represented by dynamical models. We believe unlike in computer science, memory is not a static term of information contained but is a dynamical entity that is inclusive of processes such as what to store, what information to train up and learn, and how to change, recombine, and refine representations over time. Each expert in our framework is represented as a dynamical model whose memory or representations are stored at a local level within the expert and globally as a cumulation of all the experts across a layer and across all layers. However, unlike static data these memories or representations are continually updated, improved, or recombined as needed either during online or offline processing in the inner or outer loop. Lastly, due to their statistical power, machine and deep learning techniques are used as a means to understand and optimize the non-linearity of these dynamical models.

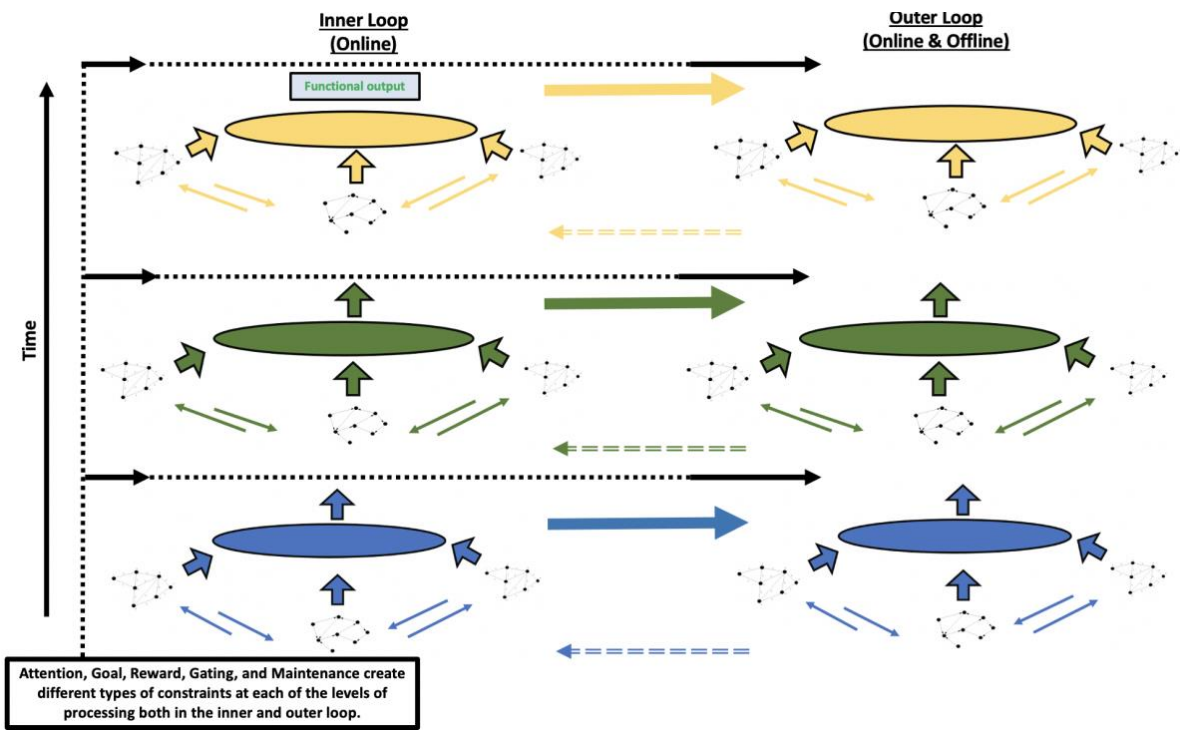

**Figure 1:** Diagram of the Ten Commandment Framework

# The Ten Commandments of Brain Function

## I) The brain uses multiple experts trained on multiple modalities which influence each other laterally

Unlike traditional computational techniques, the brain has a very rich, complex, and multi-modal environment to learn from. For example: humans do not just learn the category of a dog as a visual representation but rather a multidimensional array that includes a multitude of senses, including auditory (barking), olfactory (musky smell), somatosensory (sensation of fur) etc. These representations are each constructed by a unique expert that processes information based on special structural analysis (for example: Fourier analysis for auditory input by the auditory cortex or edge, color, and texture detection for visual input by the visual cortex). Though these different experts across modalities are optimized for their specific role, they have the ability to laterally influence one another to help fine tune representations and encourage the binding of activation across experts to the same object/category (O'Reilly & Munakata, 2000, Herd et al., 2006, Munakata et al., 2011).

This ability to have multiple experts that are each specialized for a particular type of processing allows for a better definition of inputs, loss functions, processing, and the generation of outputs. Such an architecture also allows for more explainability and transparency in the generation of outputs for a network. Furthermore, the ability to have experts laterally influence other experts allows for the creation of richer abstractions that are constrained across multidimensional modalities leading to a highly tuned local and global representation. Inspired by the Global WorkSpace Architecture (Baars 1993, 1994, 1997, 2002, & 2005) such multi-modal techniques have begun to be incorporated in deep networks such as Radford et al., 2021, Goyal et al., 2021, Bengio, 2017, and Dehaene & Kouider, 2021. Furthermore, lateral connectivity in CNN's (Pogodin et al., 2021) has also been utilized to showcase improved performance.

## II) The brain uses sparse distributed representations and bidirectional connectivity

An overarching goal of evolution seems to be to reduce uncertainty in the face of noise. Natural intelligence has to navigate noise both externally in the real world and internally in the brain. The brain employs copious amounts of noise and neurons have to learn to be selectively trained to ensure that only a specialized set of neurons are active above threshold to respond to cues (Bear et al., 2020). Though the neurons that get active are highly tuned, they are also flexible, and the failure of one or two neurons to fire in the sequence does not lead to a person not being able to perform a task (Rolls & Treves, 2011, Radford et al., 2021).

Sparse distributed representations are the technique the brain uses to ensure a multitude of different ways of categorizing an input to be active at the same time, i.e., such as across experts. The successive activation of these representations across different levels and areas of the brain is believed to be a leading driving force for the emergence of intelligent behavior (Beyeler et. al, 2017, Nair, 2021, Ahmad & Scheinkman, 2019, Ahmad & Hawkins, 2016). Furthermore, bidirectional connectivity uses these sparse distributed representations to let neural activation across numerous brain regions work together to encode complex representations and abstractions (O'Reilly et al., 2012).

Bidirectional connectivity is also essential to multiple constraint satisfaction, attention, and attractor dynamics. It allows for the creation of stable and cleaned-up representation of a noisy input by allowing a network to stabilize (O'Reilly, Munakata, Frank et al., 2012). This mechanism is however obviously further strengthened by learning dynamics as discussed below. Some artificial neural network architectures have begun to utilize these techniques such as Grewal, 2021, Hunter et al., 2021.

### III) The brain uses abstractions

Though not very much is known about the neural processes that allow the brain to make the leap to achieve so much from so little, many suspect the magic lies in the types of abstractions it creates (van Kesteren et al., 2012; Gilboa & Marlatte, 2017; van Kesteren & Meeter, 2020). The brain is an optimized system of general intelligence which is believed to be capable of learning quickly and dynamically by creating flexible knowledge structures that can be combined, recombined, and applied in new and novel ways. These knowledge structures allow organisms to ingest the world by efficiently breaking complex inputs down and encoding information into numerous small blocks or abstract mental structures. These abstractions are then internally processed to construct an internal model of the interpreted reality. Though the progression of inputs to models isn't well understood, it is widely hypothesized that the brain does so by organizing into global gradients of abstraction (Taylor et al., 2015; Mesulam, 1998; Jones & Powell, 1970).

This organization of information into global gradients of abstraction and upon which learning operations can be thought of as a continuous attractor model (Whittington, Muller, Mark et al., 2019). In this framing, as information is taken in, the attractor states stabilize into common attractor states via error driven learning, with cleaned-up stable representations of the noisy input pattern (O'Reilly et al., 2012). The stable representations that emerge employ self-organization learning to in turn build knowledge structures and systems from which complex cognition can emerge. This building up of representations across levels of a hierarchy allows for information to be scaled up from input to the pinnacle layers of cognition such as reasoning, consciousness, and other tangible behaviors of higher-level cognition. Furthermore, both structural and functional connectivity serve as important divisions that play different but interacting roles to achieve abstraction and subsequently higher-level cognition within the brain (Nair, 2021).

Though originating in neuroscience, the hypothesis of abstraction formation and attractor dynamics has been highly influential to the AI community (Mozer et al., 2018, Ren et al., 2019, Buckner, 2018, Ilin et al., 2017, Ashok et al., 2020). Though it is not fully understood to what level the abstractions formed in deep neural networks are akin to the brain, it does serve as the basis of current and future research.

## IV) The brain uses a hierarchy

It has long been hypothesized that cognitive behaviors exist on a hierarchy, where abstractions are slowly combined by increasing levels of dimensionality to arrive at higher level representations. (Botvinick, 2008; Badre & Nee, 2018; D'Mello et al., 2020). According to many neuroscientists including Taylor et al., 2015, the lowest levels of the pyramid structure represent inputs such as visual, auditory, and somatosensory sensation, whereas the highest levels represent consciousness, imagination, reasoning, and thinking.

As abstract representations are built up through the levels of the hierarchy both within the inner and outer loop, the information across regions of processing is eventually converted into generalized knowledge structures. In this hierarchy, specific brain regions or experts house the structural representations at varying degrees of complexity and dimensionality locally whereas their consolidated activation across a layer leads to the corresponding specialized functional representations. These representations can be optimized to construct actions and plans for the task at hand in the inner loop or fed into the outer loop to create even more complex high dimensional abstractions over time that can create long term goals, outcomes, and eventually world models. Furthermore this information can be used to optimize the inner loop and long term cognition in general. Each of these local and global layers are likely able to organize into generalized knowledge structures by constraint satisfaction operating at different levels of analysis.

Though current state of the art deep networks do utilize hierarchies for informational processing (Murdock et al., 2016, Qi, 2016, Kriegeskorte, 2015), to our knowledge they do not include the use of experts of the creation of multiple layers of analysis akin to the brain. We believe this slow build-up of contextualized expert knowledge gives the brain some of its cognitive power.

V) The brain has areas with specialized roles

As suggested in commandment I, the brain learns to create multimodal representations that can be constrained by multiple experts across a variety of modalities. Though deep neural networks have begun to utilize specialized models for multimodal tasks (Radford et al., 2021) or multiple heads in reinforcement learning (Schrittwieser et al., 2020, Ye et al., 2021), the brain is far more specialized in optimizing its inputs at a structural level. To put this into an example, in the visual processing domain alone, the brain has a hierarchy of specialized regions processing information.

For example: lower brain regions such as the primary visual cortex or V1 are optimized to process edges. Whereas areas such as V2 take that information and processes it for the next level of complexity. This can include determining color, texture, and background-foreground detection. Beyond these layers, the information is further specialized by being split into a 'What' and a 'Where' pathway, and the information is distributed across pathways based on type. The top of the 'What' hierarchy is the Inferior Temporal Cortex (IT) which represents complete objects, and the top of the 'Where' hierarchy include regions that guide saccades or eye movement based on constraints such as attention, maintenance, and reward. These same constraints are further imposed on lower levels of the 'Where' pathways through a top-down influence. This top-down influence is what we believe results in constraints (such attention, maintenance, and reward) to manifest in different ways across different structural experts in our framework.

VI) The brain has needs, but also goals, wants, and desires

Organisms in the natural world are goal-oriented, likely engrained from evolution itself. Humans and animals have needs that have to be met to prolong survival. When a particular need becomes activated, it is converted into a goal, which mobilizes the system to assemble actions to gain a particular kind of reward. Though behavior doesn't always have to be goal-directed, as seen in exploration or trial-and-error strategies, salient events are learned even in these instances by being contextualized into an action-outcome pairing focused on a goal which can be exploited later.

Deep neural networks on the other hand do not formulate the world into actionable objectives as humans do, because they do not face the same environmental and resource pressures. Though it could be argued deep networks have objectives they need to meet, generally their survival and health is not directly dependent on it. Humans and animals learn quickly from the world around them not only because they can form flexible multi-modal representations that they use to act in the world but also because the training signal they learn from is far more potent. For example, eating the wrong food or not attending to a predator can have dire consequences for a living organism whereas a deep neural network would get a punishment signal and restart its game. Due to this survival bias, the brain has large amounts of real-estate devoted to processing and learning negative information and the predicting away of reward through the dopamine pathway (Mollick et al., 2020, Schechtman et al., 2010, Pignatelli & Beyeler, 2019). Thus, this suggests the brain is optimized to fine-tune learning to meet survival needs, be that in the continuous attainment of relevant goals or learning from negative events within a few exposures.

Though goals are important, they are not absolutes. A particular goal can take precedence in guiding behavior, however pursuing this same goal can be maintained in the background if a different salient event directs attention to another objective. For example: If an animal is foraging for food, the goal of hunger is directing behavior to maximize food, however, if a salient event directs attention such as the location of a predator nearby, the goal of food would be maintained as a sub-goal and the new goal of escaping the predictor would dominate behavior. Sub-goals are additionally important in dividing up accomplishing a complex task into more attainable sub-components.

Furthermore, unlike deep neural networks, aside from goals, it is known that at least humans have needs, wants, and desires. We believe this is a manifestation of the motivational component of formulating rewards and the attainment of goals with respect to time. Needs can be considered actionable items that are imperative to survival, which when activated result in a goal that requires fulfillment in a short time frame. Wants, on the other hand, are needs which are not immediately survival-sensitive, and thus are not as swiftly pursued (for example: wanting a piece of chocolate). Lastly, desires are the more complex objectives that require meeting multiple sub-goals and are least time-sensitive (for example: having the desire to travel to Tuscany). These wants and desires are likely only seen in humans or maybe animals capable of more complex cognition, to drive the system to obtain new or better rewards that are not currently present or essential to survival. As far as we are aware this type of modeling has not been done in traditional deep networks and has only been done partially in other artificial neural networks (Herd et al., 2021, O'Reilly et al., 2010, O'Reilly et al., 2014, O'Reilly et al., 2020).

VII) The brain is more than just reinforcement learning

Though reinforcement learning is arguably an integral and essential part of what the brain uses to learn about its world, it is not the only mechanism used to learn (Sutton & Barto, 2018, Gariépy et la., 2014, O'Reilly et al., 2021, Doll et al., 2009, Mollick et al., 2020, Liakoni et al., 2022). Traditional deep learning networks use an actor-critic system akin to the basal ganglia and corticostriatal and mesocortical dopamine systems found in the brain (Mollick et al., 2020, Herd et al., 2021, Sutton & Barto, 2018, Schrittwieser et al., 2020, Silver et al., 2021). In the system, the actor takes inputs of the state and outputs a best action (or policy), whereas the critic evaluates the action by comparing it to a state-value function (Sutton & Barto, 2018, Li, 2017).

Though reinforcement learning is part of what the brain is doing, it does so by utilizing a myriad of complex specialized architectures. For example: the OFC is believed to be responsible for emotional and motivational value of a stimulus, the ACC is important for conflict and error monitoring, the DLPFC is important for complex higher-order informational processing, and PFC is important for integrating inputs from these areas and coordinating with basal ganglia loops to facilitate action responses (O'Reilly, 2010, Droutman et al., 2015, Bear et al., 2020, O'Reilly et al., 2012). In comparison to traditional deep networks, the brain has a much more complex hierarchical and specialized architecture that allows for the processing of different features that define rewards. These specialized regions are further driven to learn through four core mechanisms; 1) specialized architectures that build up in complexity hierarchically; 2) top-down influence and the use of constraints such as attention, maintenance, and reward that manifest in different ways across levels and experts to force the system to better its learning and outputs; 3) the use of predictive learning as a mechanism to foretell the outcome an action will have based on what the system knows; and 4) the utilization of a slew of different types of learning such as supervised, self-supervised, unsupervised, predictive, instructional, mirroring, trial and error, to train up and define its reward signal so that it can better understand the world.

We believe a grandiose benefit of using combination learning, constrained by attention, maintenance, and gating results in the utilization of different techniques that constrain the system to form different but overlapping representations of the same action-outcome. This facilitates the creation of more flexible representations that can generalize. This has been seen to some extent in robotic learning (Chen et al., 2021). Additionally, the training of reward as a means of attaining a particular goal helps contextualized learning so that different rewards can be seen differently, for example: in humans the reward of eating food when hungry, is different from the reward of buying your first home. To our knowledge this has not been attempted in deep networks, however we believe this would be immensely helpful in better contextualizing learning.

VIII) The brain uses a serial architecture or serial-parallel architecture

It is believed that decision making, or complex cognition does not operate in parallel in the human brain (Herd et al., 2021, O'Reilly et al.,2020, Hayden, 2018, Hunt et al, 2018, Herd et al., 2022 in print). The most intuitive way to think about this is that as the brain has specialized structures or experts specializing in specific components of informational processing, all experts across levels cannot be engaged all at once. Rather as information is inputted into the brain, it unfolds sequentially across multiple iterations. Brain regions in layer one of the hierarchy likely learn to process the inputs in parallel operations before serially passing along the consolidated representations to the layer two experts and so on. In this process, information is systemically processed by relevant areas in parallel before being communicated serially to create more complex abstractions, plans, and internal models. Traditional deep networks on the other hand, process multiple informational streams in parallel to maximize GPU architectures. Though they have some specialization that requires serial processing further into the model, it is quite different from the brain (Jin et al, 2020, Deng, 2011).

It is believed the serial manner of encoding information into a network and parallel processing across experts in a particular layer is also what makes the brain so well conditioned to complex decision making. A common parallel drawn to this is the think-aloud protocols of decision making (Herd et al., 2021, Herd et al., 2022 in print). In order to accomplish a complex goal-oriented task we generally divide it out into smaller serial steps and process information only relevant to those steps. For example: when planning a trip; step 1 involves assembling a list of transportation options; step 2 involves choosing which mode of transportation operates within our time constraints; step 3 involves comparing costs; and step 4 requires booking the best option (Herd et al., 2021). Interestingly enough, this same manner of serial-parallel processing also applies to learning new material. For example: when learning general relativity as a novice, step 1 involves learning simple concepts like What is a planet? What is a Sun? What is a Solar System?; step 2 takes these informational inputs to build more complex hypotheses and abstractions such as what is an orbit or what keeps these objects from colliding into each other, i.e. gravity; step 3 addresses the concept of space, time, and space-time ; and step 4 addresses what is general relativity.

This means of hierarchically building up complex representations as information unfolds serially over time, we believe is an essential ingredient of higher order cognition. Furthermore, the process of being able to internally query representation at different levels of abstractions, using statements such as when, how, and why allows us to better direct our learning to form causal inferences that better refine and consolidate representations and direct attention. Some research in neuroscience has further suggested that animal recordings have found the use of a serial-parallel architecture when evaluating options (Hunt et al, 2018). Lastly, though a serial-parallel architecture is optimal for learning, it is believed that the brain can optimize over time by using a parallel architecture in well-practiced tasks (Herd et al., 2021, Herd et al., 2022 in print).

## IX) The brain thinks and reasons

We believe a significant distinguishing factor between human and machine performance comes from the ability to think and reason. We find it likely that young children, much like current artificial intelligence tools, develop simple abstractions and internal models of the world, making them only as good as what they train on. However, over the course of development children can take what they know, recombine and refine it to build increasingly complex adult representations and models of the world. We believe this is possible as humans have the ability to think, or wander through their representational space. Humans, like artificial intelligence tools, also have computational or physical constraints. For instance, they cannot continuously act in the world, such as, endlessly searching for food when hungry or water when thirsty will eventually deplete all their energy and result in death. Instead, humans formulate complex plans mentally and then choose which actions they should execute in the world. This ability to mentally simulate the world allows us to combine not only representations that we have learned should be paired together, but also new combinations of representations.

Much like a neural network that can use its off-policy to influence its on-policy, humans too can go offline, to combine representations to influence online behavior. In our framework, when the system chooses to engage in thinking, it likely goes offline by reducing the effect of constraints for the inner loop sensory mechanisms (vision, sensory, auditory etc), i.e, less attention, maintenance, and gating is directed to sensory inputs. However, the constraints upon the inner loop lower-level sensory areas can easily be reengaged in response to a salient stimuli in the world that gets above threshold. The system, by reducing focus on the inputs of the world, can instead direct computational processing to the abstractions already formed in higher levels of the system. When the system is thinking in a time-sensitive context about the current task it is solving, only the abstractions present in the inner loop are available. This includes the abstractions that are task relevant as well as any abstractions from the outer loop that have previously been refined to influence inner loop behavior. Active processing in the outer loop much like the sensory areas of the inner loop will likely be disengaged by reducing the effect of the constraints. Contrairly, when the system chooses to engage in time-insensitive thinking or goal-insensitive thinking, active processing likely only occurs in the outer loop, by reducing focus on the inner loop or current environment. However, this operation of limiting processing to only the inner or outer loop is not perfect. When attention and other constraints are not imposed correctly in the inner loop, the system would reactivate the outer loop and bring in new abstractions which may not be task relevant and cause interference for timely processing. For example: thinking about lunch when working on solving a homework problem.

It should also be noted that mind-wandering, i.e. searching through representational space without a strong influence of a goal is possible due to the imperfection of constraints being imposed. When a particular task is time relevant but current abstractions cannot converge on a solution, constraints such as attention, gating, and maintenance cannot impose appropriate dominance on the system. This results in the system wandering between both inner and outer loop representations. For example: when a deadline is looming and the solution to a complex problem cannot be found, the constraints imposed over the inner loop fatigue and causes the system to wander through representational space in both loops unbounded till the system can recover and the constraints correctly can be imposed again. The cause for the constraints to weaken would occur from computational fatigue due to exponential expenditure of resources with a lack of convergence of the system.

These mechanisms of mental simulation, memory recall, and likely mind-wander are all a depiction of the default mode network (Schacter et al., 2012, Zhou & Lei, 2018). Some research additionally suggests this same default mode network is also active during sleep (Horovitz et al., 2009, Sämann et al., 2011, De Havas et al., 2012). The ability for the system to wander offline or entirely disengaged during sleep, could be the necessary mechanism that allows for the refinement and consolidation of information, and the flexible combination of abstractions that leads to more complex behaviors such as imagination, creativity, problem solving, and generalization (Rasch & Born, 2013). During mind-wandering, we believe the effect of the constraints on all levels of the inner loop is substantially reduced. Whereas, during sleep on the other hand, we believe only the higher levels of the outer loop are available, which is why dreams operate in the most abstract domains. Lastly, we believe that when a task is less time sensitive, only the outer loop is engaged. For example: when planning how to spend a weekend, the system can now combine different and new combinations of representations without a task-relevant tie.

Reasoning, a sub-component of thinking, is a mechanism that allows the system to better direct and refine its search in representational space. We believe reasoning is the formulation of causal relationships between representations. Though it is likely that early learning is driven by simplistic means such as similarity and dissimilarity, pattern matching, and reinforcement learning as more complex mechanisms like predictive learning formulate, the system likely uses a process of internal querying as relevant to the goal to contextualize the world and formulate casual relationships. These internal querying processes would use statements such as "how", "why", and "when" as a means for hypothesis testing action-outcome pairs, changes in the world, and the relevance of actions to the goal. As far as we are aware such mechanisms for thinking and reasoning have not been considered or implemented in deep neural networks. However, examples of hard-coded reasoning outer loops using SAT solvers and imagination monte carlo tree search roll outs have been implemented (Ye et al., 202, Chen et al., 2019, Schrittwieser et al., 2020). Other biologically inspired models include Russin et al., 2020 and Russin et al., 2021.

X) The brain operates in collaboration with other brains

There has long been a saying that *"there is strength in numbers"*. This is equally true when it comes to building better ideas, or in this case smarter civilizations. People don't learn in isolation, outside of their internal architectures, they learn both from rich multi-modal environments and from others (Bassett & Mattar, 2017). Humans have specialized mirror neurons to mimic the actions taken by individuals in the world around them. This learning by imitation also propagates the same pathways to fire, when we are thinking about enacting those actions in the world as well as when we actually act or are instructed to act (Heyes, 2010). Furthermore, in addition to cooperation dynamics, the role of competition also plays an important role in driving pressures on the system to learn representations better and faster (Pinto et al., 2017, Baker et al., 2019,

Sanchez, 2017). We believe the incorporation of a social aspect to the learning process will result in the creation of a better and faster system.

Recently, the use of multi-agent learning has been incorporated into deep learning networks with great success. The most notable examples are Open AI's hide and seek agents, Deepminds XLAND agents (Stooke et al., 2021, Baker et al., 2019), who have demonstrated a range of complex emergent behaviors. In robotics, human training and human-machine teaming as well as competition vs. cooperation experiments have demonstrated faster and more efficient learning (Chen et al., 2019, Pinto et al., 2017).

## Assembling the Ten Commandment Framework

As discussed above, our framework proposes a means to house the ten commandments into a system that can be mathematically and computationally executed. The commandments build upon each other to result in the emergence of behavior that is more powerful than its individual components. However, it should be explicitly noted that the aim of this framework is not to suggest an absolute solution to the problem, but instead inspire future architectures and discussions that could too build upon these pillars of strength.

As described above, the framework consists of two hierarchical loops; an inner loop that corresponds to processing the task at hand and a slower more integrative outer loop which assembles information on a longer time frame (O'Reilly et al., 2020, Cushman & Morris, 2015, Schneider & Logan, 2006). To assemble the inner loop first, each layer in the hierarchy is constructed of experts. As shown in Figure 2a and 2b, three expert networks *($x_1$, $y_1$, $z_1$)* work together to assemble the output of the first level $g_1(x) = [a_1(t), b_1(t), c_1(t)]$. Though any number of experts could be used, they must obey three defining rules; 1) each expert is designed to accomplish only one type of processing (for example: visual processing, auditory analysis, etc.), 2) the experts in a layer must be laterally connected, and 3) the information output from each expert is in a common functional format that can be expressed throughout the architecture (i.e., a global communication language).

The three experts *($x_1$, $y_1$, $z_1$)* as shown in Figure 2b, are each in themselves data driven non-autonomous dynamical systems $\dot{x} = f(x,t,u;\beta)$. Each expert would have its own set of variables that would describe its dynamical system where $f$ represents the dynamics (i.e, given a state how does the system change at the next time step), $x$ represents the state, $t$ represents continuous time, $u$ represents control inputs, and $\beta$ represents parameters (Brin, & Stuck, 2002). For each dynamical model, $x$, $u$, and $\beta$ would be vectors where, $x$ is the state of the system and $\beta$ is the different multimodal inputs into the system. These inputs could be brightness, color, and contrast for the visual system, sharpness, or volume for the auditory system, and texture, roughness, and temperature for the somatosensory system. At higher levels of analysis, described below these inputs could include predicted value, error signals, motor actions, proposed plans, or any factors that change the dynamics of the system. The vector $u$ would be the different controls that are imposed on the system, including attention, maintenance, and gating of information. These controls exert influence over the system, and we believe would be learned by the system over time through the overarching $f(x)$ goal function as well as the reinforcement signal. As the network receives rewards for action-outcome pairs, attention would be an emergent control that attends to relevant salient stimuli in the scene, which would influence the maintenance and gating of information in memory. Lastly, $\dot{x}$ would be the differential equation (or a vector field) which would represent the dynamics of the system.

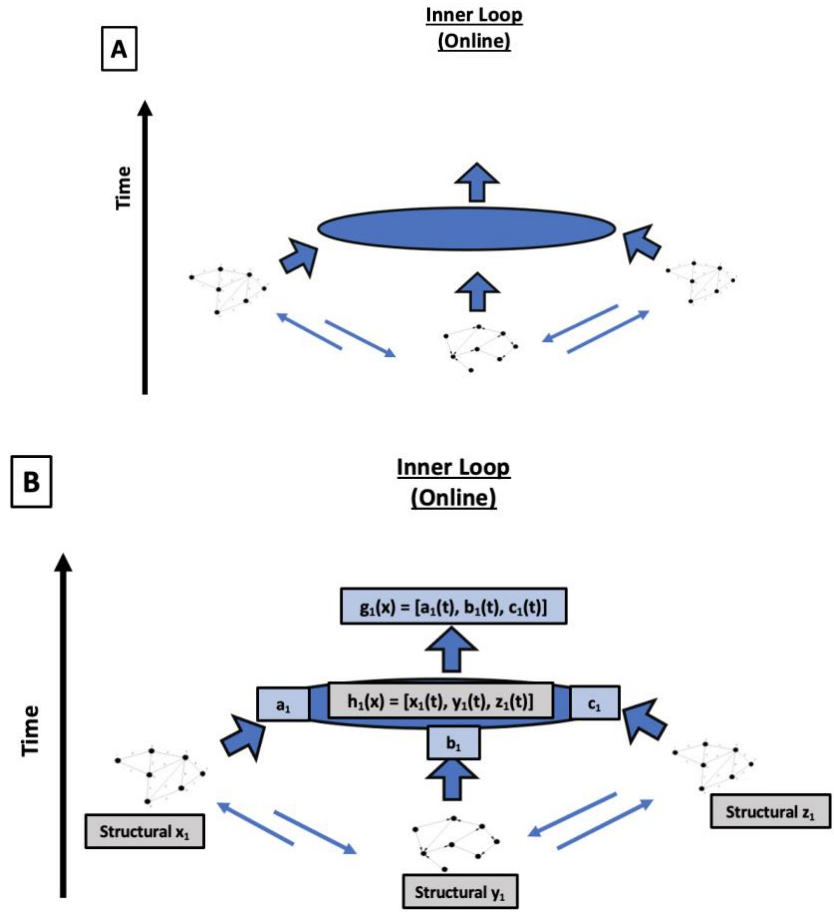

**Figure 2:** a) represents the framework for the inner loop with one layer, and b) represents the mathematical explanation for the inner loop with one layer given below.

Agreeably, by using a data driven dynamical modeling approach, this framework steps away from traditional machine and deep learning architectures. However, we believe that current deep and machine learning networks are essential to the success of finding interpretable solutions to understanding the dynamics of the system due to their statistical power. The data fed into the system would differ by expert, but the assumptions that the data comes from an autonomous system would hold true. The data for each expert would be assembled into an initial non-value problem with the intention of the network learning model propagation. The overarching benefit of the statistical power of these deep learning techniques should be helpful in determining unknown $f$ functions, shifting through the chaos, as well as finding coordinate transformations or a new coordinate system such that the transformation makes nonlinear dynamics look linear. Techniques such as sparse regression or Koopman analysis have already begun to be used (Brunton et al., 2016 & Brunton et al., 2016).

As machine learning has mastered the statistical space, these methodologies would be prime tools in combating the many challenges that arise with non-linear dynamical models. Challenges such as unknown and nonlinear $f$ functions, uncertainty and chaos, and latent and hidden variables could be rephrased as optimization problems, where the machine learning models could be optimized for the data (Brunton et al., 2022, Lusch et al., 2018). In our framework, machine learning models could be locally implemented to understand dynamics of each of the experts, and globally in determining the overall output of the framework, i.e., the collective system of all dynamical model experts used to create feedback control. This would entail performing optimization across experts based on the inputs and actively manipulating parts of the system in real time to convert parameters into actuators that transform the model into a feedback control loop. Though this methodology would propose new challenges in the development of next generation artificial intelligence tools, we believe it will lead to more transparent and generalizable networks. Through the use of a framework with specialized experts, inputs and outputs can better be constrained and expert specific transparency tools can be developed. Debugging and retraining experts would be simpler than current networks which have more complex input, output, and integration structures. Additionally, as dynamical models understand the underlying dynamics of the system, i.e, the why and how a system changes, we believe they should be much easier to adapt to new and unseen situations. It has been shown that dynamical models generalize for parameters that have not been tested, for example Newton's second law of motion (Brunton et al., 2016). We believe these generalizable properties should hold for this collective framework of experts as well.

When represented in matrix notation, eigenvalues reveal what a dynamical system will do over time, or better serve as solutions to these systems. Systems are commonly some combination of positive, negative, and imaginary eigenvalues. Negative values signify decrease and decay or a stable system, positive values signify increase and explosion, or instability, and imaginary values signify oscillations or error and conflict. It could be possible these negative eigenvalues could refer to a decrease in random activity and the formation of stable attractor dynamics across networks, whereas positive eigenvalues could refer to an increase in random activity and the formation of chaotic and unstable attractor representations. Lastly, the role of imaginary eigenvalues could perhaps extend to errors in activation in some parts of the system which could cause incorrect or delayed recall. This could result in activity not causing the correct attractor state to become completely active, resulting in scenarios such as misremembering the name of a category or incorrectly recalling past information.

A secondary technique and use of eigenvalues that we believe is key to understanding abstractions, attractor states, and creating a global language across the framework is graph theory. As machines and deep learning methods can aid in providing an analysis of the underlying system as well as finding data driven solutions for converting the nonlinear space into linear. We believe graph theory and its sub-component spectral graph theory can be used in determining and understanding knowledge components within these systems. As discussed above, dynamical systems like the brain are continuous attractor models that organize into global gradients that can be continually updated, and recombined. As information is input, these attractor states stabilize into common attractor states which use learning to create stable representations that over time can build knowledge structures and systems (O'Reilly et al., 2012, Nair, 2021). Distributed representations, occurring through bidirectional and lateral connectivity are key to ensuring this can occur across multiple regions or experts and drive the emergence of intelligent behavior (Nair, 2021, O'Reilly et al., 2012). In our framework, each expert will be equipped with the needed lateral and bidirectional connections both within the layer and across multiple layers.

As shown in Figure 2a and 2b, three structural expert networks *($x_1$, $y_1$, $z_1$)* work together to process their respective tasks. This structural information would then be converted into a functional representation $g_1(x) = [a_1(t), b_1(t), c_1(t)]$, which is a holistic picture of the most important components from each of the experts. For example: if each of the three experts are trained on visual *($x_1$)*, auditory *($y_1$)*, and somatosensory components *($z_1$)* of a category dog, the collective structural output of the system would be the sum of their dynamics, represented by $h_1(x) = [x_1(t), y_1(t), z_1(t)]$. The functional representation, on the other hand, would encompass the most relevant features across experts for that category and provide a global language that can be communicated across the framework. The structural information of the experts *$x_1$, $y_1$, $z_1$* would be outputted in a common format (*$a_1$, $b_1$, $c_1$*) that can be communicated across the layer to constrain and influence the other experts, i.e., the lateral influence across experts would happen on these common language representations. For example: constraining the visual of a dog (obtained from expert *$x_1$* and now represented by a global communicable output *$a_1$)*, with the auditory sounds of barking *(*obtained from expert *$y_1$* and now represented by a global communicable output *$b_1$)*, and the somatosensory representation of fur *(*obtained from expert *$z_1$* and now represented by a global communicable output *$c_1$)*. Through lateral connectivity, these common language outputs (*$a_1$, $b_1$, $c_1$*) influence the creation of a holistic picture of the experts' individual processing and dynamics by creating complex multimodal representation of information. These common language outputs are additionally in themselves non-autonomous dynamical systems which contributed to the final functional output for the layer $g_1(x) = [a_1(t), b_1(t), c_1(t)]$.

In the brain, it has been found that neural oscillations are a linear superposition of eigenmodes (Raj et al., 2020). Abdelnour et al., 2018, demonstrated that structural connectivity and resting state functional connectivity are related through a Laplacian-eigen structure. The researchers examined this hypothesis by using eigen decomposition of the structural graph to predict the relationship between structural and functional connectivity. This model of eigenvector full functional connectivity was compared and verified against healthy functional and structural human data, as well as nonlinear neural simulations. Through this, the authors were able to verify that Laplacian eigenvectors can predict functional networks from independent component analysis at a group structural level. Using these findings, we believe that using inter-regional couplings weighted by anatomical activity, i.e, structural experts, would influence neural node activity to show the same functional results within our framework (Nair, 2021). Thus, we propose that spectral graph theory can be used to isolate clusters and constrain neural clusters of abstractions based on eigen-relationships both within experts and across structural and functional relationships. As hypothesized in Nair (2020), by visualizing the brain as a network where neurons serve as nodes linked by weighted connectivity, graph theory can be used to determine algebraic connectivity. Furthermore, spectral graph theory analysis upon this network can divide connectivity into its smallest subcomponents resulting in abstraction clusters which have high internal connectivity and low external connectivity. These clusters could suggest how information is being retained and manipulated within and across structural experts and the framework as a whole. Transparency tools further can be built to understand how abstractions link, update, combine, and recombine together to get a better understanding of the system.

The next steps to assembling this framework are shown in Figure 3 a and 3 b. Experts can be organized into layers that are hierarchically built for serially processing of information over time. Though three layers, each with a set number of experts have been shown here, $n$ number of layers with $n$ or $m$ number experts can exist across layers. The importance of having a serial and hierarchical flow of information is that at each layer different levels of processing occur which inform the next set of experts that can further construct increasingly complex representations of the input information to determine appropriate outputs. Figure 3 c and 3 d, further demonstrate and build up the mathematical explanation given above for the single layer for layer two and three of the inner loop. The mathematical processing across each of these layers of experts remains the same as layer one, till processing completes and an output is executed. As information is processed by the first layer of experts, the functional output is then used to inform the next set of structural experts at level two, who perform a more complex level of analysis upon these abstractions. The functional outputs produced in the first layer are converted back into structural inputs through a laplacian eigen conversion to pass on to the structural experts of the second layer. Similarly, these structural experts perform specific processing and output a common language representation which is constrained laterally and outputted as a final functional representation of the layer to pass on to the next set of experts in layer three. Layer three, in our example, is the final step of the framework. It similarly performs the final step of processing and executes an output into the world.

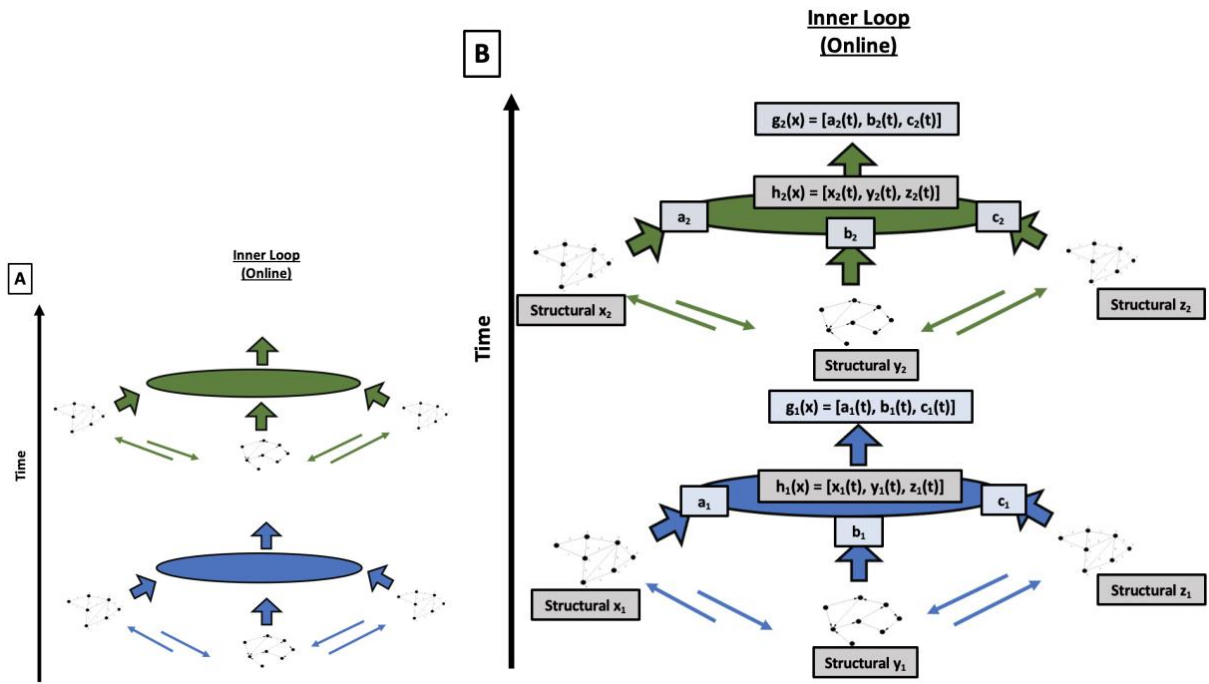

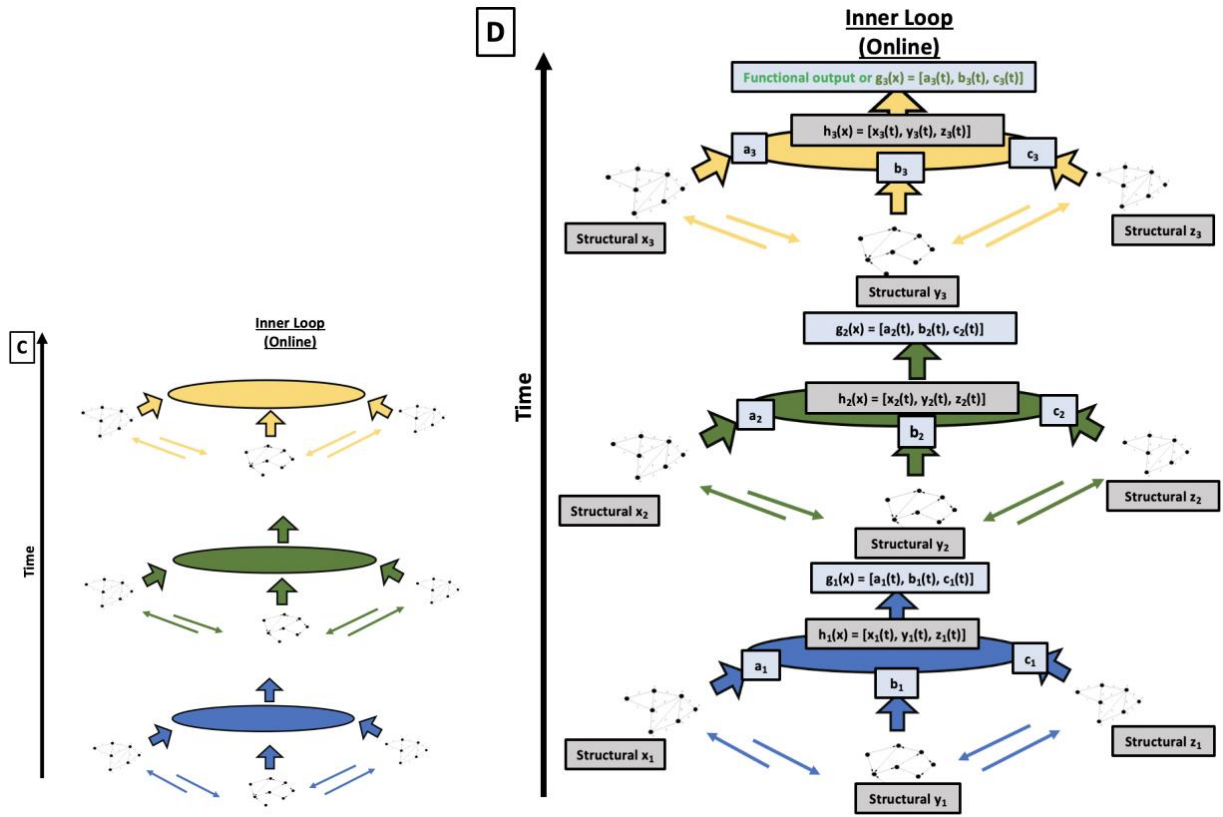

**Figure 3:** a) represents the framework for the inner loop expanded out to two layers, b) represents the mathematical explanation for the inner loop expanded out to two layers given above, c) represents the framework for the inner loop expanded out to three layers, and d) represents the mathematical explanation for the inner loop expanded out to two layers given above.

To elaborate, if the first layers of experts are performing visual ($x_1$), auditory ($y_1$), and somatosensory ($z_1$) analysis of a dog. The functional output $g_1(x) = [a_1(t), b_1(t), c_1(t)]$ would be converted back into structural inputs that would then be inputted to the relevant experts at the second layer. The second layer of processing would consolidate the environmental inputs from layer one to achieve a higher level of processing. This could include experts such as scene integration ($x_2$), language ($y_2$), and motor action and planning ($z_2$). Similar to layer one, the structural inputs would be consolidated as $h_2(x) = [x_2(t), y_2(t), z_2(t)]$, and converted into a global common language $a_2, b_2, c_2$ upon which lateral connectivity can operate to output the functional representation of the layer $g_2(x) = [a_2(t), b_2(t), c_2(t)]$. Similarly, the functional outputs from layer two would be converted back into structural inputs and inputted respectfully to the relevant experts in layer three. This layer would consolidate the information from layer two to achieve a final level of abstraction processing for determining the output. This could include experts such as value prediction ($x_3$), value evaluation and error ($y_3$), and motor execution ($z_3$). These experts would integrate information from abstractions of lower layers to execute an action that maximizes the goal, or the overarching *f(x)* function. The goal would be representative of an active need that the framework has to achieve, such as reaching a charging station for a robot. As previously computed in past layers, the consolidated structural inputs would be consolidated as $h_3(x) = [x_3(t), y_3(t), z_3(t)]$, and converted into a global common language $a_3, b_3, c_3$ upon which lateral connectivity can operate to determine the final functional representation that would generate the output.

Additionally, as all layers have bi-directional connectivity, layer three can exert top-down influence upon layer two to redirect experts to encode new or specific information to generate better outputs at the next time step. Layer two can similarly exert a top-down influence upon layer one during or after execution to gather better information. Furthermore, if this information gathered leads to better action-outcome mappings, the relevant objects in the scene are remembered, or gated and maintained and attention is directed to them in the future. These same attention, maintenance, and gating mechanisms serve as the constraints that are imposed upon the dynamical systems of experts. Each of these constraints would manifest differently at different levels of abstraction, for instance at lower levels of the framework it could direct scene search, whereas at higher levels of the framework it could direct semantic search for reward values associated with a particular action-outcome pairing in the past.

It should be noted that each level of the framework doesn't build on complexity from subcortical to cortical brain regions but rather corresponds to a buildup in hierarchical processing of abstractions. For instance, experts in layer one of the hierarchy used for sensorimotor processing would include lower level layers such as primary visual cortex (V1) as well as higher level cortical areas such as Inferior Temporal Cortex (IT). Whereas experts in layer two would include regions such as supplementary motor area (SMA) which proposed internally generated planning of movement, and premotor cortex which is responsible for aspects of motor control including the preparation for movement. These areas would process the information from the scene inputted from layer one to plan motor outputs to be evaluated in layer three. Building upon the information from layer two, layer three would construct a final functional output of the system. Layer three could include brain regions such as the primary motor cortex which generates neural impulses to execute movement, and the prefrontal cortex (PFC) which integrates different reward values and chooses which actions to go (act on) and which actions to no-go on (not act on), via the basal ganglia. This operation, akin to a mixture of experts approach, would biologically combine experts from lower levels to higher levels of abstraction processing. Furthermore, though this framework may not be an entirely accurate account of brain function, it is believed higher level brain areas operate and process increasingly complex abstractions from lower regions (Nair, 2020, Richards et al., 2019). This can be seen in human development, as children build abstractions of their environments through the sensorimotor regions that they can then use in higher regions of cortex to enable more complex cognitive planning. As abstractions of the environment improve, so does higher level planning (Markovits & Vachon, 1990).

After assembling the inner loop, we next move to assembling the outer loop, the slower integrative system. As shown in Figure 4 a, 4b, and 4c, each layer and expert in the inner loop corresponds to a mirrored layer and expert in the outer loop. The intention of having two separate identical hierarchical serial processing loops is to create a secondary system that can accumulate and assimilate relevant information from the inner loop to achieve more complex and longer time-scale cognition. Thus, each gated representation from each of the inner loop layers is passed to the corresponding outer loop layer where refinement, updating, and recombination of representations can occur. As mentioned earlier, even though three layers with three experts each are shown for both the inner and outer loop, there can be $n$ number of layers and $n$ or $m$ number of experts per layer in the framework. The only constraints that exist are the number of layers and the number of experts per layer in the inner loop have to match those in the corresponding outer loop. Though this architecture might seem contrary and duplicative, we believe it can result in a framework that is more generalizable, explainable, and embodies properties of high-level cognitive capabilities. Additionally, from a neuroscience perspective, duplicative brain areas do not exist in this two-loop hierarchical format. However, we believe the same brain regions perform both the role of the inner and outer loop. By dividing the framework into duplicates, we believe we can create a more interpretable framework of representations as it pertains to the current task and how they are refined over time and learning.

As assembled, the inner loop processes the current state of the world and formulates action-outcome representations that maximize reward for a particular or multiple goals. The inner loop actively engages, processes, and acts in the world, constraining representations across experts with lateral connectivity and formulating controlled feedback through top down and bidirectional connectivity between layers. The action-outcome representations that lead to reward are maintained and gated, thus training up relevant weights and representations through the system. Though the type of learning used has not been explicitly specified, we believe either backpropagation or a dopamine based learning signal could be used. This would be subject to experimentation depending on which signal leads to better performance in the system. However, we believe the dynamical and multifaceted properties of the dopamine system would likely lead to superior performance. These gated representations from the inner loop for the current task are passed to the respective outer looped mimicked experts and layers. For instance, $h_1(x) = [x_1(t), y_1(t), z_1(t)]$ and $g_1(x) = [a_1(t), b_1(t), c_1(t)]$, would be inputted to the corresponding layer one outer loop layer and experts, to eventually lead to $H_1(x) = [X_1(t), Y_1(t), Z_1(t)]$ and $G_1(x) = [A_1(t), B_1(t), C_1(t)]$, which are better consolidated outer loop representations of the inputs created over time. This transfer of gated representations from the inner to the outer loop happens for all tasks and at all relevant gated timesteps while the system is interacting with the real world.

The role of the outer loop is to take these representations and consolidate information by refining, recombining, and updating these gated representations in a separate loop that does not directly interact with the real world. For instance, the inner loop may use different types of learning to correctly identify action-outcome pairings in a task. This could be inclusive of instructional learning and unsupervised learning strategies used at different time steps that result in obtaining the same goal but using a different combination of actions. The gated representations for both types of learning would then be sent to the outer loop, whose experts would work to refine and recombine the representations in abstract space and propose a new consolidated strategy that is a combination of the best elements of the previous strategies enacted by the model in the real world. These new combinations of representations would be exerted upon the inner loop by the outer loop using bidirectional and top-down influence when the framework is put in that same situation in a future time step. For instance, $G_3(x) = [A_3(t), B_3(t), C_3(t)]$ exerting influence over the inner loop layer three experts, $g_3(x) = [a_3(t), b_3(t), c_3(t)]$, that determine final functional output.

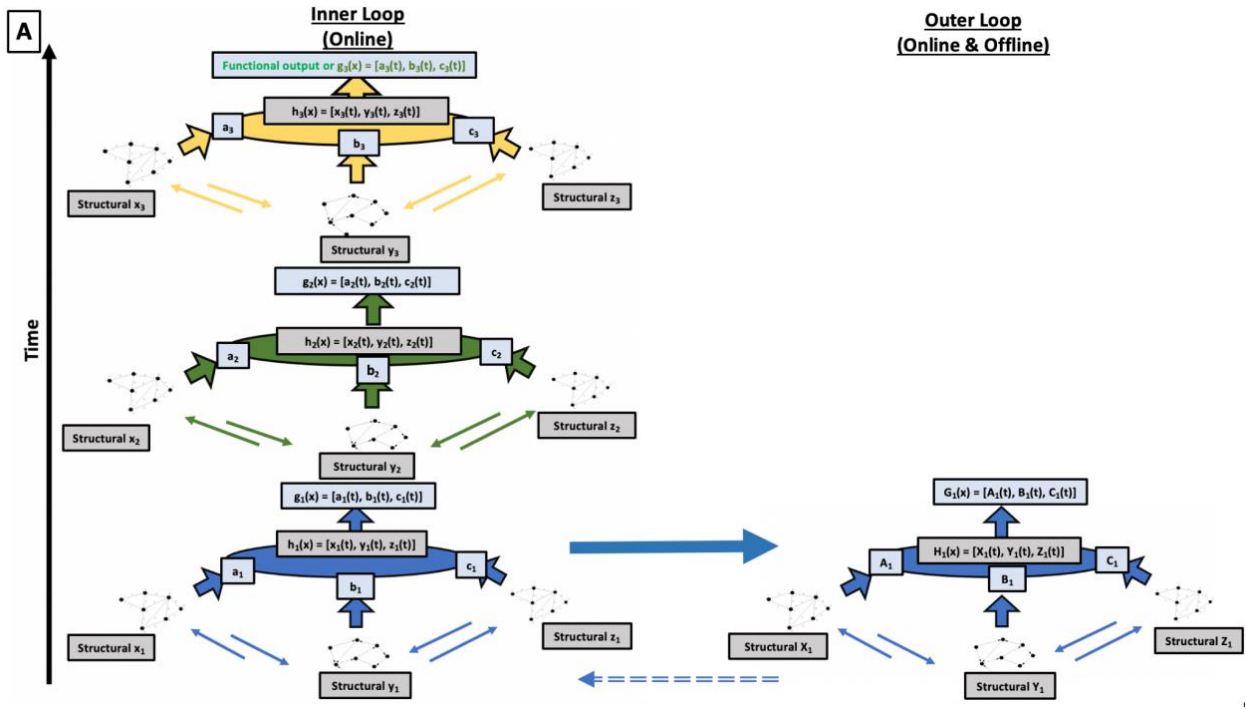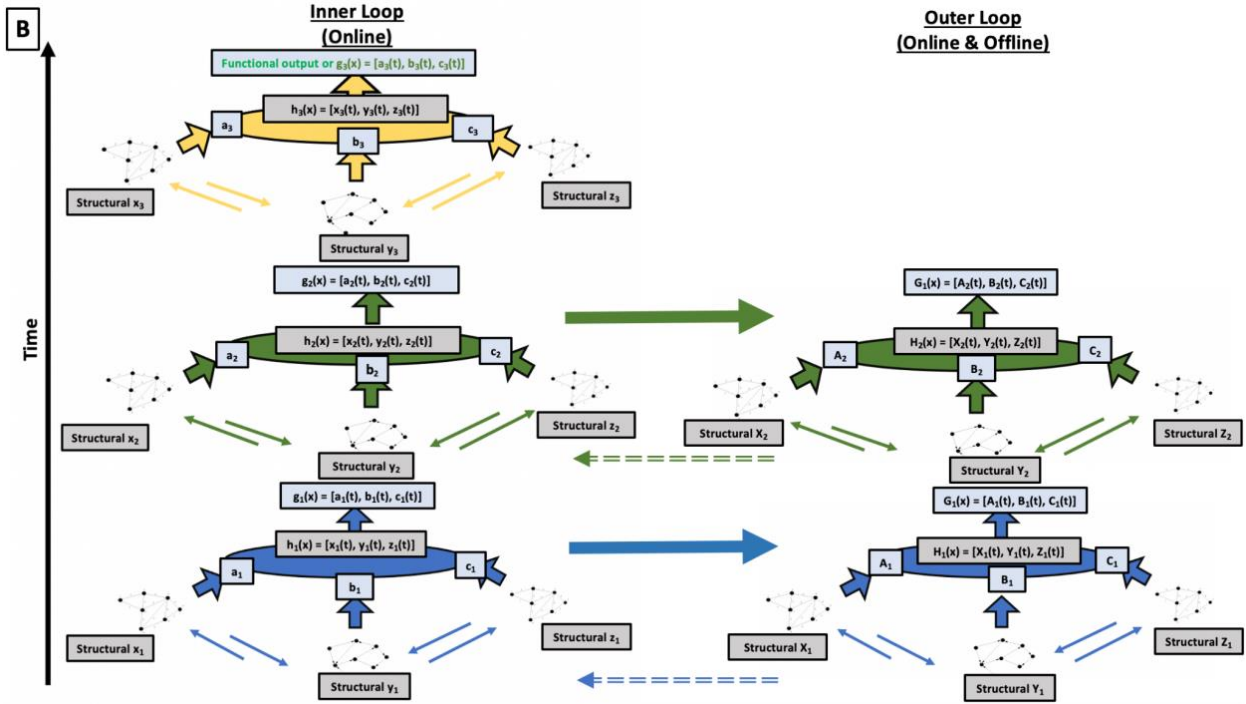

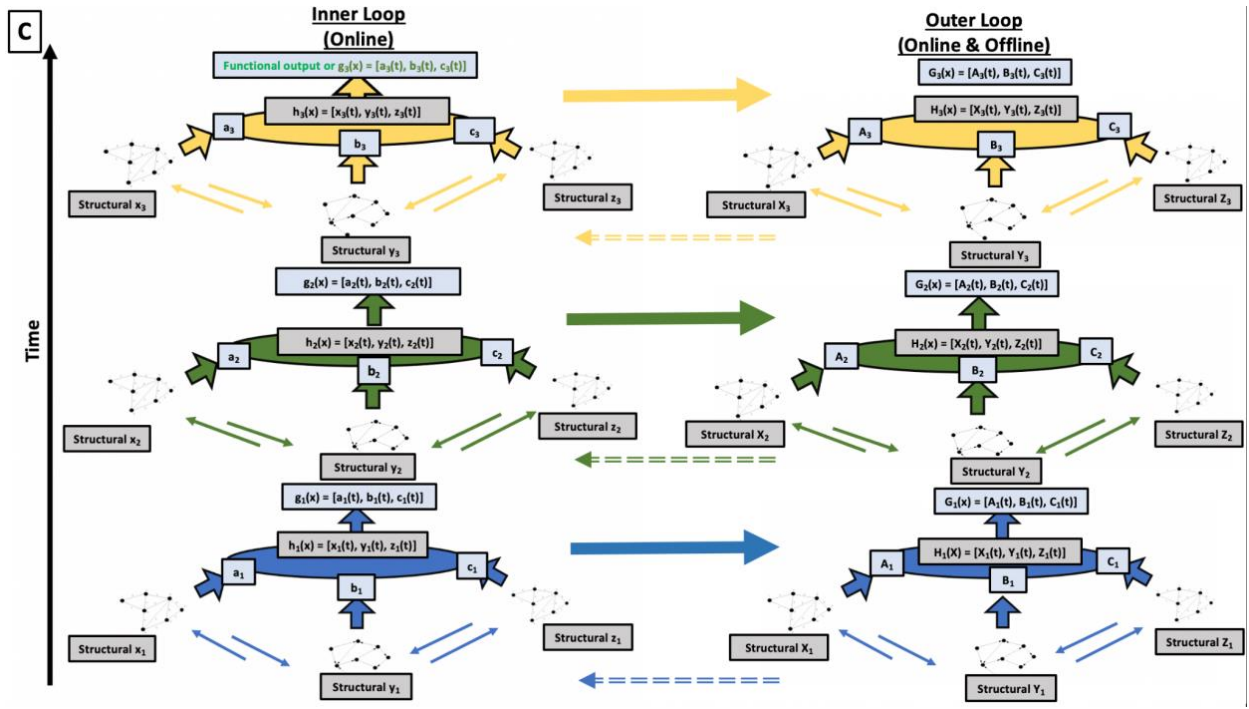

**Figure 4:** a) represents the mathematical explanation of the framework for the inner and outer loop expanded out to one layer, b) represents the mathematical explanation of the framework for the inner and outer loop expanded out to two layers, c) represents the mathematical explanation of the framework for the inner and outer loop expanded out to three layers.

Another importance of the outer loop is its contribution to thinking and reasoning, i.e., taking what the system knows, and recombining and refining this information to build new and better representations and models of the world. In humans, it is believed this is possible through the ability to think or wander through representational space. As discussed above, this ability to combine information offline, allows for natural intelligence to manage computational and physical constraints by only acting in the real world with purpose. In time-sensitive contexts, the inner loop would solve a task in two ways. Firstly, the higher-level layers of the hierarchy could direct lower-level systems to collect additional information from the environment that can be continually processed online. Secondly, the lower-level areas of the inner loop would be disengaged by reducing the effect of the constraints such as attention, and computational processing would be redirected to higher level regions. This strategy would involve using only existing representations from the inner loop as well as representations from the outer loop that have been previously refined and are communicated via top down and bi-directional control. This type of processing would include situations like how to determine the quickest route to the store in your hometown.

On the other hand, in time-insensitive contexts, the amount of influence exerted by the goal would depend on how pressing achieving the goal would be. In situations where a solution is needed to be determined in a day's time, the goal would exert a stronger influence over the framework than a situation where a solution needed to be determined in a week's time. In these situations, active processing would likely occur entirely in the outer loop, by reducing attention on both the inner loop and current environment. During this processing, the system would wander through representations that had been gated and stored in the outer loop. The representations would be directed to be combined in new and novel ways through reasoning, reflection, and internal querying. Reflection would allow for the framework to compare predicted outcomes with rewards received in the context of achieving a goal. These representations would be examined using an internal querying process which would hypothesize how, and why the existing dynamics differed from achieving the goal, and reasoning would be enlisted to form new causal relationships that could drive the system to a new higher reward value. These new representations would be maintained as plans to be imposed on the inner loop at relevant time points. We believe that these outer loop processes can be further expanded out over longer time scales, akin to mind wandering and dreaming.

As in this framework, it should also be noted that the brain uses the same mechanisms for remembering the past and simulating the future (Schacter et al., 2012). We believe through the use of dynamical models of representations and the construction of an outer loop assigned to consolidating and combining representations for both reflecting on past events and simulating new future events, we can embody these same properties of cognition. Additionally, a benefit we believe should arise from this framework is serial and parallel processing akin to the brain. When the inner loop is deciphering how to succeed in a new task, it inputs information to be processed serially and hierarchically across loops. However, once the framework has mastered the task, it can process multiple information streams within the two loops in parallel, for instance, automatic assignment of value to multiple sensory stimuli or action. Lastly, though not explicitly mentioned, we believe that by the use of noise and the above-mentioned commandments, an emergent property of the system would be the utilization of sparse distributed representations and attractor dynamics. This could be further analyzed using spectral graph theory, which would determine clusters that are strongly connected internally and weakly connected externally.

In conclusion, we will root this framework into a real-world example utilizing a robot trained to perform the coffee making task (Tsai et al., 2010). As paralleled in a human in the introductory text, the objective of this task is for the robot to correctly identify all the objects in the scene such as coffee, water, parts of the coffee machine and assemble the right combination of steps needed to make coffee. The experts in layer one of the inner loop would intake information of the environment such as visual inputs of objects, the sounds different objects make, the somatosensory feel of the objects, and the pretrained motor actions needed to reach for these objects. Through layer one of the framework, the robot would interact with its environment in a goal oriented manner, with the overarching *f(x)* goal function guiding the system. This information would then be sent to the second layer of the framework, at which stage more complex processing such as scene integration, plan proposal, and motor action planning can happen across these individual laterally connected experts. This stage would integrate the sensorimotor information coming from level one, integrate motor feedback, and propose a plan of action. Through bi-directional connectivity and hinging on the processing happening at layer two, the experts at this level can exert a top-down influence upon the lower layer experts to collect new information to formulate better action plans. Additionally, this could include encoding new or specific visual, auditory, or somatosensory information about objects in the scene. In early stages of training (such as trial and error learning) or if the framework is under immense time pressure, we believe the processing in level two could lead to a direct-action output executed by level three without additional processing. However, as this strategy would ignore higher level processing, and instead use simple strategies such as past reinforcement learning values, and similarity and dissimilarity analysis, it is likely this approach would lead to a lower level of reward and a lower probability of achieving the current goal. However, such strategies can be useful in a speed vs. accuracy tradeoff or in simple situations that don't require extensive processing (Herd et al., 2022, in print).

As representations are refined in level two, the abstractions formulated can then be sent to layer three for final processing. This could include processes such as reward value evaluation and outcome prediction, and motor action execution. This final stage of processing would involve making predictions of action-outcome values, integrating errors from past actions, and assembling a motor action plan for execution. After the model executes the actions and receives an outcome from the environment, the representations involved in the execution of the outcome can be either trained up for meeting the goal, i.e., reward or trained down for not meeting the goal, i.e., punishment. The information that is gated through this process at each level is then transferred to the corresponding outer loop layer and expert. The first layer of the outer loop would perform higher level processing of the gated inputs. This would include fine tuning pairings of sensorimotor information and making comparisons with similar or dissimilar representations used previously that matched that goal. The information would then be sent to level two of the outer loop which would consolidate all components present in a scene and assign reward value to objects and motor plans so that attention can be imposed by the inner loop faster at a later time point. If the framework is unable to consolidate these representations in a novel way or consolidate actions for faster execution, top down influence would be exerted back to the first level of the outer loop hierarchy. This would force the sensorimotor outputs in the outer loop to perform new processing on their representations or retain the request to exert influence via attention on the inner loop to attend to that information when put in a similar situation in the future.

The information processed in level two is finally sent to the highest level of the outer loop, level three. At this final stage of processing, the representations of motor plans are compared with previous combinations and outcomes through similarity-dissimilarity or reward comparison. If the system has the ability to go offline and engage in thinking, the framework would be allowed to wander in representational space. This would include three behaviors: 1) reflection of why a particular action did not lead to an outcome in the past, and engaging in new combinations of outcomes that would lead to a higher or intended reward-goal matched state; 2) through the use of a language model like GPT-3, the model could generate simple sentences to examine what the goal is, why the internal state did not lead to reward, and how a new combination of actions could lead to a higher goal-matched reward state in the future. The use of the partial differential equations to understand the underlying dynamics of the system is key for the framework to make these associations; 3) through the evaluation of representations and the formulation of internal queries to determine how and why actions lead to outcomes, the system can formulate causal dynamics to understand input-output and action-outcome relationships. Through this process of mixture of experts, hierarchical processing unfolding serially over time, and the utilization of two loops we believe robotics and artificial intelligence networks can faster learn and generalize to tasks in a more explainable manner. This same unfolding of processing could similarly be applied to other examples like self-driving cars, as well as robotic swarms to enable the social components of intelligence that was highlighted in commandment ten.

## Conclusions

In conclusion, though, the last wave of AI development saw great triumphs. There is still much that separates these AI tools from natural intelligence, especially in comparison to the exceptional general learners that humans are. Thus, in this paper, we identify components of brain function upon which we believe human intelligence is systematically and hierarchically built. These components are referred to as the the ten commandments. We believe these commandments work collectively in a system to serve as the essential ingredients that lead to the emergence of higher-order cognition and intelligence. A defining feature of this framework is that it does not draw its strength from an individual commandment, but rather from the system as a whole, that combines all components to emerge stronger than its parts. Though this framework works to serve as an all-encompassing theory of intelligence, the aim of this paper serves equally to identify fundamental principles that can help overcome current AI challenges as well as inform the next wave of intelligent models and fully autonomous or embodied AGI. These ten commandments are lastly embodied into a framework that uses a mixture of experts' approach involving the use of dynamical systems and two loops of cognitive control. We present this work not as an absolute solution but as inspiration to future architectures and discussions that could too build upon pillars of strength seen in brain function.